\documentclass[10pt, journal,twoside]{IEEEtran}
\usepackage{graphicx}
\usepackage{subfigure}
\usepackage{amsmath}
\usepackage{amssymb}
\usepackage{cases}
\usepackage{times} 
\usepackage[font=footnotesize,labelfont=bf]{caption}
\usepackage{caption}
\usepackage{algorithm}
\usepackage{algorithmic}
\usepackage{color}
\usepackage{multirow}
\usepackage{algorithm, algorithmic}

\makeatletter
  \newcommand\figcaption{\def\@captype{figure}\caption}
  \newcommand\tabcaption{\def\@captype{table}\caption}
\makeatother
\usepackage{setspace}
\usepackage[utf8]{inputenc}
\usepackage[T1]{fontenc}
\usepackage{array} 
\newcolumntype{C}[1]{>{\centering}p{#1}}
\usepackage{multicol}  
\usepackage{multirow} 

\begin{document}

\title{Machine Unlearning in Contrastive Learning}
\author{Zixin Wang and Kongyang Chen}

\IEEEtitleabstractindextext{
\begin{abstract}
Machine unlearning is a complex process that necessitates the model to diminish the influence of the training data while keeping the loss of accuracy to a minimum. Despite the numerous studies on machine unlearning in recent years, the majority of them have primarily focused on supervised learning models, leaving research on contrastive learning models relatively underexplored. With the conviction that self-supervised learning harbors a promising potential, surpassing or rivaling that of supervised learning, we set out to investigate methods for machine unlearning centered around contrastive learning models. In this study, we introduce a novel gradient constraint-based approach for training the model to effectively achieve machine unlearning. Our method only necessitates a minimal number of training epochs and the identification of the data slated for unlearning. Remarkably, our approach demonstrates proficient performance not only on contrastive learning models but also on supervised learning models, showcasing its versatility and adaptability in various learning paradigms.
\end{abstract}
}
\maketitle
\IEEEdisplaynontitleabstractindextext
\IEEEpeerreviewmaketitle

\section{Introduction}
In contemporary society, artificial intelligence (AI) has become increasingly pervasive, with numerous AI applications leveraging machine learning models. AI has permeated various aspects of human society, encompassing learning, work, and daily life. However, model privacy has emerged as a significant concern, as models may inadvertently expose individual users' privacy. For example, membership inference attacks capitalize on the discrepancies between training and non-training data predictions to infer whether specific data were utilized for model training, thereby exposing privacy risks. Additional challenges to model privacy and security include backdoor attacks and model adversarial attacks.

Moreover, privacy regulations such as the European General Data Protection Regulation (GDPR)\cite{GFPR} afford users the right to request the deletion of their personal data from learning models, a component of the "right to be forgotten." The Protection of Personal Information Act (APPI)\cite{APPI} and Canada's proposed Consumer Privacy Protection Act (CPPA)\cite{CPPA}, both mandate the deletion of private information.Erasing data from learning models is a challenging task requiring the selective reversal of the learning process. In the absence of targeted methods, the sole option is retraining the model, a costly and feasible approach only when the original data remains accessible. As a remedial measure, researchers like Cao and Yang, and Bourtoule et al.\cite{bourtoule2021machine} have proposed machine unlearning methods. These techniques partially reverse the learning process, facilitating the retrospective deletion of specific data points, mitigating privacy breaches, and addressing user deletion requests. Yan et al.\cite{yan2022arcane} and Ga et al.\cite{wu2022puma} introduced an approximate unlearning method, achieving effects akin to retraining with minimal additional training.

Nonetheless, these methods exhibit limitations, chiefly their dependence on supervised learning for machine unlearning. Research on contrastive and self-supervised learning remains relatively limited. In this study, we present a novel gradient penalty-based machine unlearning method, enabling approximate unlearning by modifying the loss during model training. This approach requires minimal training to effectuate machine unlearning for designated data while ensuring that the model can forget specified data without a substantial loss in accuracy (generally within 10\%). Our method is simple, efficient, and highly adaptable, demonstrating commendable performance on both supervised and contrastive learning models.

\section{Related Work}
\textbf{Contrast learning and contrast learning models:} Contrastive learning and contrastive learning models: Contrastive learning is referred to as self-supervised learning in some literature, while others call it unsupervised learning. Self-supervised learning is a form of unsupervised learning, and there is no formal distinction between the two in existing literature, so both terms can be used interchangeably. The core idea behind contrastive learning is quite simple. Suppose we have three images A, B, and C, with two images belonging to the same category and one image belonging to a different category. In supervised learning for classification problems, we want the model to recognize A and B as the same category defined by our labels, while C is another category. However, for contrastive learning, we only need the model to know that A and B are similar, belonging to the same category, and that C is different from A and B, belonging to a different category. Contrastive learning models do not need to know the specific category of A and B. That is, contrastive learning does not need to know the true labels of each image, only who is similar to whom and who is dissimilar.

Assuming A, B, and C all pass through a network, we obtain features A', B', and C' for the three images. We hope that contrastive learning can bring A' and B' closer in the feature space and keep them away from C'. In other words, the goal of contrastive learning is to have all similar objects in adjacent regions in the feature space while dissimilar objects are in non-adjacent regions. 

Chen Ting et al.\cite{chen2020simclr} proposed a contrastive learning framework called SimCLR for learning visual representations. It uses contrastive learning to train neural networks with unlabeled data. The core steps include: generating positive sample pairs through data augmentation, feature extraction, projection to a low-dimensional space, and calculating contrastive loss. The training objective is to bring views from the same original image closer in the low-dimensional space while keeping views from different images apart. After training, the encoder can be used to extract features for new images for downstream tasks.

Kaiming He et al.\cite{He_2020_CVPR} proposed MoCo. Compared to SimCLR, MoCo introduces a momentum encoder and a memory queue. The core steps include: generating positive sample pairs through data augmentation, feature extraction, encoding negative samples with the momentum encoder, storing historical negative sample representations in the memory queue, and calculating contrastive loss. The training objective is similar to SimCLR, and after training, the base encoder can be used to extract features for new images for downstream tasks.

GRILL et al.\cite{grill2020bootstrap} proposed BYOL, which is used for learning visual representations. Unlike SimCLR and MoCo, BYOL does not require negative sample pairs and only uses positive sample pairs generated through data augmentation for learning. Compared to SimCLR and MoCo, the main feature of BYOL is that it does not require negative sample pairs and relies solely on positive sample pairs for learning. After training, the online encoder can be used to extract features for new images for downstream tasks.

\textbf{Member inference attacks and model overfitting: }As artificial intelligence applications become increasingly popular, the security of models has begun to gain attention. Many studies have shown that models that learn too quickly may be prone to overfitting\cite{overfitting1}\cite{overfitting2}. Although the models do not store the data used for training, the training data still exists within the model in another form. For example, when we use training data and non-training data to let the model make predictions and obtain the probability values of the prediction output, we find that the probability distribution of the prediction output for the training data is extremely extreme, with the top1 or top3 highest prediction probabilities being tens or hundreds of times higher than other categories. In contrast, the prediction probability distribution for non-training data is not as extreme, and when compared to the training data, the probability distribution is relatively flat.

Membership inference attacks\cite{mia} are methods used to infer whether some or a batch of data belongs to the model's training data. In supervised learning, we only need the target model's prediction output for the data we want to infer, and then we can use these outputs to generate training data for a binary classifier, which can recognize whether the data is a member data. In contrastive learning, we can exploit the differences in cosine similarity values between training data and non-training data when augmenting data to train a binary classifier to distinguish between training and non-training data. The emergence of membership inference attacks poses significant challenges to the privacy and security of models.

Laws and regulations such as GDPR and CCPA stipulate that companies using AI models must adhere to the following requirements: data subjects, i.e., the individual sources of data, have the right to request that companies using their personal information to train AI models delete their personal data without delay. However, deleting data from a model, specifically the data used to train the model, is not as straightforward as traditionally thought, where simply deleting the data from a database would suffice. In fact, once the model training is complete, there is no need to save the training data, and the model does not memorize the training data. The role of the training data is to help adjust the internal parameters of the model so that the parameters are tuned to the correct position. Therefore, making a model forget data has become a very meaningful research direction.

\textbf{Machine Unlearning: }The premise of machine unlearning\cite{MUsurvey} is to enable a model to completely forget the influence of specified data, with the main idea being to directly remove the data to be forgotten from the entire model training process. Based on this concept, two directions for machine unlearning have emerged: complete unlearning and approximate unlearning.

Complete unlearning entails retraining the entire model. SISA\cite{Machine_unlearning} (Sliceable Incremental and Selective Aggregation) is an early method for machine unlearning. The core idea of SISA is to split the original dataset into multiple independent subsets, which do not share information during training. These independent subsets are then incrementally trained by slicing and partitioning the data. Incremental training implies that the model trains on each data slice and updates the model parameters after each training session, allowing the model to gradually adapt to all sliced datasets. Finally, an ensemble method is used to combine these models. One ensemble approach calculates the output vector for each model, averages these output vectors, and selects the maximum value from the mean vector as the final classification result. The advantage of this method is that it can achieve complete unlearning of the data to be forgotten. However, its drawbacks are that it requires adopting a specific framework, which may not be compatible with existing models and training frameworks used by most companies, and the loss of model accuracy due to the ensemble method can be substantial, greatly impacting the model's performance.

Due to these drawbacks, approximate unlearning has been considered as an alternative. Approximate unlearning does not require the model to achieve the same effect as with non-trained data; it only needs to be close. Moreover, it can be achieved by training the existing model for a small number of iterations. This approach not only saves a significant amount of training resources but also preserves the model's original performance to the greatest extent. PUMA\cite{wu2022puma} (Private Update via Model Approximation) is a method for implementing approximate unlearning. Its goal is to remove the influence of training data while maintaining minimal changes in model performance. PUMA achieves this primarily by generating synthetic data and fine-tuning the model using this data. However, one drawback is that generating perturbation data with activation values similar to the forgotten samples in the model can be challenging. This process may require complex optimization methods, such as gradient matching, thereby increasing computational complexity.

The method proposed in this paper is also an approximate unlearning approach. However, the key difference is that our method can be applied to both supervised learning and contrastive learning models, making it a versatile solution for a wider range of applications.

\section{Gradient penalty-based unlearning method}
In this section, we will specify the details of our unlearning method, and the source of inspiration for our method
\subsection{gradient penalty:}
Our approach is inspired by WGAN\cite{WGAN}, in which the generator needs to compute gradient\_penalty as a loss to ensure the stability of the model training. The gradient penalty ensures that the gradient of the discriminator remains appropriate during training by interpolating between the real and generated samples and requiring the gradient of the interpolated points to be close to 1 in magnitude. This gradient penalty term will be added to the loss function of the discriminator to ensure the stability of the training process. By using the gradient penalty, WGAN-GP has better stability in training the game process between the generator and the discriminator, and reduces the risk of gradient disappearance and pattern collapse problems. The following is the flow chart of the algorithm for the penalty term.
\begin{algorithm}
	\caption{Compute Gradient Penalty for WGAN GP}
	\begin{algorithmic}[1]
		\STATE \textbf{Input:} Discriminator $D$, non-trained samples $nontrained\_samples$, trained samples $trained\_samples$
		
		\STATE // Compute random weight term for interpolation between non-trained and trained samples
		\STATE $alpha = Tensor(np.random.random((nontrained\_samples.size(0), 1, 1, 1)), device=nontrained\_samples.device)$
		
		\STATE // Get random interpolation between non-trained and trained samples
		\STATE $interpolates = (alpha * nontrained\_samples + ((1 - alpha) * trained\_samples)).requires\_grad\_(True)$
		
		\STATE // Calculate the discriminator's predictions for interpolates
		\STATE $d\_interpolates = D(interpolates)$
		
		\STATE // Create fake tensor filled with 1.0, with the same shape as the discriminator's output, no gradients required
		\STATE $fake = Variable(Tensor(nontrained\_samples.shape[0], d\_interpolates.shape[1], device=nontrained\_samples.device).fill_(1.0), requires\_grad=False)$
		
		\STATE // Get gradient w.r.t. interpolates
		\STATE $gradients = autograd.grad(outputs=d\_interpolates, inputs=interpolates, grad\_outputs=fake, create\_graph=True, retain\_graph=True, only\_inputs=True)[0]$
		
		\STATE // Reshape gradients to (size, -1)
		\STATE $gradients = gradients.view(gradients.size(0), -1)$
		
		\STATE // Compute gradient penalty
		\STATE $gradient\_penalty = ((gradients.norm(2, dim=1)) ^ 2).mean()$
		
		\RETURN $gradient\_penalty$
	\end{algorithmic}
\end{algorithm}

Through our investigation, we have discovered that by treating real samples as trained samples and fake samples as non-trained samples, the gradient penalty loss can bring the model's prediction confidence for both trained and non-trained data closer together. This, in turn, renders the trained and non-trained data indistinguishable. The core component of this function hinges on generating interpolated data between trained and non-member data, followed by computing the respective gradients and calculating the penalty value. Below is the formula for calculating the penalty value.
\begin{equation}
	\text{gradient\_penalty} = \frac{1}{N} \sum_{i=1}^{N} \left(\left\| \nabla_{\mathbf{z}_i} D(\mathbf{z}_i) \right\|_2 - 1\right)^2
\end{equation}

However, this approach presents a notable limitation. While it successfully brings the model's prediction confidence for both member and non-member data closer together, it does so by mutually converging the predictions. In other words, the predictions for non-trained data are also altered, eventually causing the trained and non-trained data to aggregate at a central point between the two. Although this process reduces the probability distribution of the model's output prediction confidence for trained data, it concurrently increases the output prediction confidence for non-member data. This outcome is not desirable; ideally, the model should treat predictions for trained data as if it were untrained, while maintaining the non-trained data predictions unchanged.

\begin{figure}[!htb]
	\centering
	\begin{minipage}{.5\textwidth}
		\centering
		\includegraphics[width=0.8\textwidth]{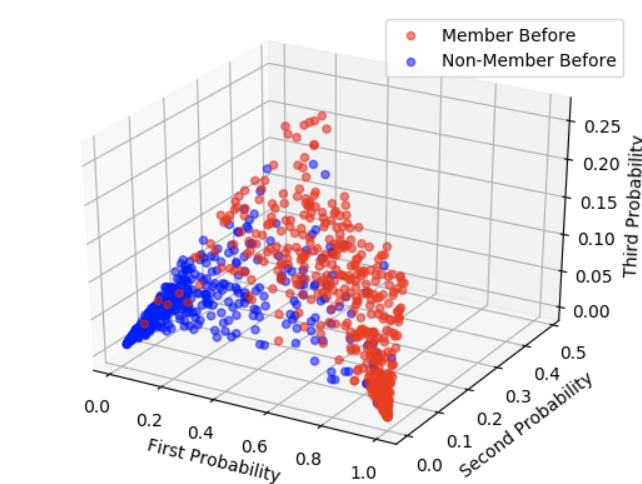}
		\caption{Use gradient penalty before}
		\label{fig:MU_bef}
	\end{minipage}\hfill
	\begin{minipage}{.5\textwidth}
		\centering
		\includegraphics[width=0.8\textwidth]{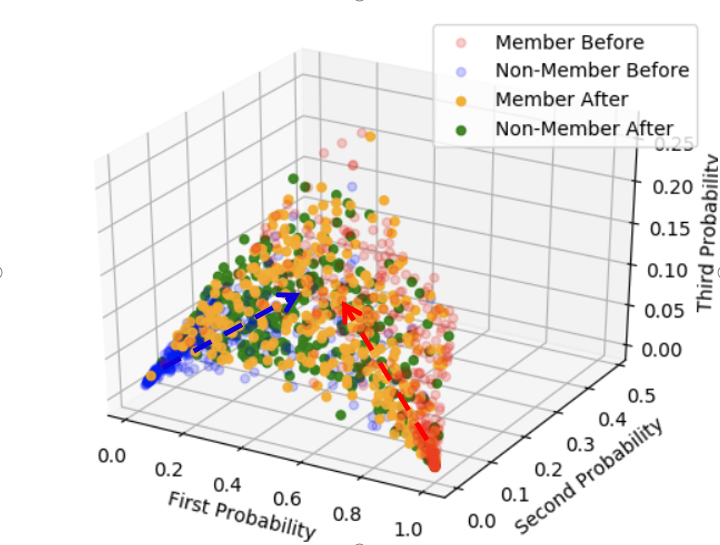}
		\caption{Use gradient penalty after}
		\label{fig:MU_af}
	\end{minipage}
\end{figure}

\subsection{Our Design Objectives:}
Due to the aforementioned issue, our primary focus in subsequent research is to identify a loss function or a specific technique capable of reducing the model's prediction confidence for data. Although L2 regularization can mitigate the extent of overfitting in the model, it fails to alleviate the disparity between member and non-member data. However, since the penalty term in WGAN can reduce the difference in prediction outputs between trained and non-trained data, we propose a combination of the two approaches. By calculating the average L2 norm of the model's output for member data as a loss component and combining it with the penalty term and the member data training loss, we can create a composite loss function for limited model training. This approach enables the targeted unlearning of specific data.

With this understanding, we establish the foundation for our study on contrastive learning for machine unlearning, focusing primarily on two objectives. First, to enable the model to defend against membership inference attacks, the model's output prediction confidence for member and non-member data should be nearly identical or indistinguishable. Second, to achieve data unlearning, the model's prediction confidence for data should be relatively low, or the uncertainty should be considerably high.

\subsection{Our Method}
To accomplish these two objectives, we employ the WGAN gradient penalty term as a loss function for machine unlearning training, ensuring that the model's prediction confidence for member and non-member data is nearly indistinguishable. Subsequently, we calculate the L2 norm of the model's encoder prediction output for member data as a loss function for machine unlearning training, aiming to reduce the model's prediction confidence for data. Finally, we incorporate the model's training loss for member data as a constraint term during the machine unlearning training process to maintain the model's accuracy.

Our proposed method is capable of rendering member and non-member data predictions indistinguishable while maintaining a minimal loss in model accuracy (at most approximately 10\%). This approach is applicable to both contrastive learning models and supervised models. For contrastive learning models, we first extract the model's encoder, then perform unlearning based on the encoder's prediction output for the data. Our method requires only a simple modification to the training loss function, and demands a relatively low number of training epochs (approximately 10). The general form of our loss function is as follows:

\begin{equation}
\mathcal{L} = \alpha \cdot L_{\text{MEMtrain}} + \beta \cdot L_{\text{GP}} + \gamma \cdot L_{\text{Norm}}
\end{equation}

The loss function is composed of three distinct components. The first component corresponds to the training loss of the data to be forgotten within the model. This element primarily serves to prevent model collapse during the unlearning process. In our experiments, omitting this component resulted in significant losses in model accuracy. The second component is the gradient constraint term from WGAN, which, in its original form, stabilizes model training by preventing gradient explosion and vanishing. However, our research has discovered that this function can also facilitate convergence in prediction output confidence for both member and non-member data. The third component involves computing the L2 norm of the model's prediction output, aimed at reducing the prediction confidence. Although the second component can induce convergence in prediction confidence, it tends to elevate the confidence for non-member data predictions. Therefore, we strive to lower the model's prediction confidence for all data.

While the aforementioned method is feasible and exhibits satisfactory performance, the presence of three components in the loss function unavoidably necessitates hyperparameter adjustments to accommodate various unlearning scenarios across different datasets and models. Conversely, this also provides flexibility, as the method can achieve desired results by tweaking hyperparameters. The most pressing issue, however, is the need for the introduction of non-training data, with a distribution similar to the training data, to compute the penalty loss for GP. This requirement increases the cost of unlearning. Consequently, to address this issue and simplify the loss function, we have developed an optimized method. Following optimization, our training loss requires only the training loss for the data to be forgotten, combined with our optimized gradient penalty term, to achieve performance comparable to the previously mentioned method.

\begin{equation}
	\mathcal{L} = \alpha \cdot L_{\text{MEMtrain}} + \beta \cdot L_{\text{MEMGP}}
\end{equation}

The following describes the algorithmic flowchart for the gradient penalty term. In the optimized gradient penalty term, there is no longer a need for additional non-member data. Our method aims to directly apply gradient penalties to member data. We discovered through experimentation that by directly penalizing the gradients of member data, the prediction output confidence for member data approaches that of non-member data. As a result, there is no longer a need to employ L2 regularization to reduce the overall prediction confidence.

\begin{algorithm}
	\caption{Compute Gradient Penalty Trained for WGAN GP}
	\begin{algorithmic}[1]
		\STATE \textbf{Input:} Discriminator $D$, trained samples $trained\_samples$
		
		\STATE // Makes trained\_samples require gradients
		\STATE $trained\_samples.requires\_grad\_$(True)
		
		\STATE // Calculates the discriminator's predictions for trained samples
		\STATE $d\_trained\_samples = D(trained\_samples)$
		
		\STATE // Create fake tensor filled with 1.0, with the same shape as the discriminator's output, no gradients required
		\STATE $fake = Variable(Tensor(trained\_samples.shape[0], $ \\ $ d\_trained\_samples.shape[1], device=trained\_samples.device).fill_(1.0), requires\_grad=False)$
		
		\STATE // Get gradient w.r.t. trained\_samples
		\STATE $gradients = autograd.grad(outputs=d\_trained\_samples, inputs=trained\_samples, grad\_outputs=fake, create\_graph=True, retain\_graph=True, only\_inputs=True)[0]$
		
		\STATE // Reshape gradients to (size, -1)
		\STATE $gradients = gradients.view(gradients.size(0), -1)$
		
		\STATE // Compute gradient penalty
		\STATE $gradient\_penalty = ((gradients.norm(2, dim=1)) - 1^2).mean()$
		
		\RETURN $gradient\_penalty$
	\end{algorithmic}
\end{algorithm}

\section{Model Unlearning Review}
According to our research, we discovered that overfitting and machine unlearning are more easily detected in the encoders of contrastive learning models. Our findings show that successful membership inference attacks can be performed on the encoders of contrastive learning models by directly extracting them for generic supervised membership inference. The EncoderMI\cite{EncoderMI} study, which inspired our research, employs cosine similarity to perform membership inference attacks. Despite the absence of labels during contrastive learning training, the learned representations exhibit high prediction confidence. As a result, successful membership inference attacks can be conducted by treating the encoder as a supervised learning model.

To ensure the feasibility of our approach, we compared our method to the one presented in EncoderMI. The comparison revealed that the accuracy of the two methods is comparable, indicating similar performance. Since we aimed to evaluate machine unlearning in contrastive learning models, we introduced a new variable to assess whether the model had truly forgotten. This variable measures the cosine similarity of the model's data augmentations. To compute this, we generated two augmented data samples and calculated the positive cosine similarity loss using these samples as positive training data. The original data of the other samples in the batch were used as negative samples to calculate negative cosine similarity, bringing similar-origin data closer and distancing other data. This is the fundamental principle of contrastive learning, and by computing the cosine similarity of similar-origin data, we can determine if the model has genuinely forgotten.

In EncoderMI, the authors conducted membership inference attacks by exploiting the difference in output cosine similarities between member and non-member data. Theoretically, the cosine similarity calculated using the model's output for similar-origin augmented member data should be very high, approaching 1, while the cosine similarity for non-member data should be lower. We leveraged this difference to assess the effectiveness of our unlearning method. If our method truly induces unlearning, the cosine similarity of similar-origin augmented data for forgotten samples should be very close to that of non-member data.

We further investigated the changes in training and non-training data before and after unlearning using a multi-dimensional combination approach. By considering the top-1 prediction confidence, training loss, and cosine similarity, we created a three-dimensional space to visualize the changes in data distribution. For supervised learning unlearning evaluation, we employed membership inference attacks and t-SNE visualization to assess the effectiveness of unlearning. Additionally, we examined the distribution of training and non-training data before and after unlearning by plotting the top-3 prediction confidence values in a spatial diagram.

To evaluate the performance of the classifier in membership inference attacks, we used AUC, precision, accuracy, and recall metrics. We observed the effect of machine unlearning by comparing the distribution of training and non-training data in two-dimensional t-SNE scatter plots before and after unlearning.

\section{Experiments}
\subsection{Experimental:}
\textbf{Setup Data Set:}
We conducted experiments on the SVHN, CIFAR10, CIFAR100 data sets. The distribution of the datasets is shown in Table.\ref{tab:datasets}

\begin{table*}[!htb]
	\centering
	\begin{tabular}{|l|l|l|l|l|}
		\hline
		Datasets & Shape       & Classes & Number of training data & Number of testing data \\ 
		\hline
		CIFAR-10  & 32x32x3     & 10      & 50,000    & 10,000    \\
		CIFAR-100 & 32x32x3     & 100     & 50,000    & 10,000    \\
		SVHN      & 32x32x3     & 10      & 73,257    & 26,032    \\ 
		\hline
	\end{tabular}
	\caption{Datasets description}
	\label{tab:datasets}
\end{table*}

\textbf{Experimental Details:}
We implemented the series of attacks described above using PyTorch in Python 3.7. Our computational resources included 4 NVIDIA V100 GPUs. To control variables, we conducted experiments using a combination of 10,000 training data and 10,000 test data samples. During contrastive learning model training, we trained the model for 1,600 epochs to induce overfitting, using the Adam optimizer with a learning rate of 0.01.

Common methods for evaluating encoder performance include linear evaluation and weighted KNN evaluation. Linear evaluation measures the quality of feature representations extracted by the encoder when trained with a linear model. Weighted KNN evaluation involves comparing feature representations' cosine similarity and classifying them using a weighted voting k-nearest neighbors method. During the training process, we monitored performance using weighted KNN evaluation and tested the final performance with linear evaluation.

For the experiments, the baseline amount of data to be forgotten was 2,000 out of 10,000 training data samples, which corresponds to 20\% of the data. We conducted additional comparative experiments by varying the number of forgotten data samples. The supervised learning model used a ResNet architecture, while the contrastive learning model employed MoCo. The number of epochs required for unlearning training was 10. We also conducted ablation experiments to demonstrate the importance of loss selection in our method.

\textbf{Evaluation Metrics:}
Membership inference attacks primarily evaluate classifiers; therefore, our evaluation metrics include accuracy, precision, recall, and AUC.

Accuracy: Accuracy is the proportion of samples that the classification model correctly predicts relative to the total number of samples. It is calculated using the following formula:

$
\text{Accuracy} = \frac{\text{True Positives} + \text{True Negatives}}{\text{True Positives} + \text{False Positives} + \text{True Negatives} + \text{False Negatives}}
$

Accuracy is suitable for balanced classes; however, in imbalanced class situations, it may not accurately reflect model performance.

Precision: Precision is the proportion of true positive samples among all samples predicted as positive by the model. It is calculated using the following formula:

$
\text{Precision} = \frac{\text{True Positives}}{\text{True Positives} + \text{False Positives}}
$

Precision reflects the reliability of the model when predicting positive classes.

Recall: Recall is the proportion of true positive samples that the model correctly predicts as positive among all actual positive samples. It is calculated using the following formula:

$
\text{Recall} = \frac{\text{True Positives}}{\text{True Positives} + \text{False Negatives}}
$

Recall reflects the extent to which the model covers the detection of positive classes.

AUC (Area Under Curve): AUC represents the area under the Receiver Operating Characteristic (ROC) curve. The ROC curve is drawn based on the true positive rate (TPR) and false positive rate (FPR) at different thresholds. AUC values range from 0 to 1, with a perfect classifier having an AUC of 1 and a random classifier having an AUC of approximately 0.5. AUC is a comprehensive performance metric that can reflect the model's classification ability in imbalanced class situations.

\subsection{Experimental Results:}
We first use our machine unlearning method for the self-supervised comparison learning model, and then we use the method of that ENcoderMI paper to perform membership inference attacks on the model, and then record the change in the success rate of the membership attacks before and after performing machine unlearning, as shown in the following table.
 
\begin{table*}[!htb]
	\centering
	
	\begin{tabular}{|c|c|c|c|c|c|c|c|c|}
		\hline
		Dataset & \multicolumn{4}{c|}{Before Unlearning} & \multicolumn{4}{c|}{After Unlearning} \\
		\cline{2-9}
		& model\_acc & mia\_acc & mia\_rec & mia\_pre & model\_acc & mia\_acc & mia\_rec & mia\_pre \\
		\hline
		cifar10 & 70\% & 87\% & 87\% & 88\% & 60\% & 51\% & 52\% & 52\% \\
		\hline
		cifar100 & 32\% & 92\% & 92\% & 92\% & 25\% & 51\% & 51\% & 51\% \\
		\hline
		svhn & 76\% & 86\% & 87\% & 86\% & 73\% & 51\% & 51\% & 51\% \\
		\hline
	\end{tabular}
	\caption{Contrast Machine unlearning performance before and after Unlearning}
\end{table*}

From the above table, we can see that our method is able to be member inference attack completely invalid, while it can keep the accuracy loss of the model not too high to some extent. To further demonstrate the feasibility of our method, we experimented with different models for moco, simclr and byol, and the experimental results are shown in the following table.

\begin{table}[!htb]
	\centering
	
	\begin{tabular}{|l|c|c|c|c|}
		\hline
		Model   & ACC\_bef & ACC\_af & MIA\_bef & MIA\_af \\ \hline
		MoCo    & 70\%      & 60\%    & 90\%     & 50\%    \\ \hline
		SimCLR  & 66\%      & 60\%    & 70\%     & 50\%    \\ \hline
		BYOL    & 55\%      & 49\%    & 70\%     & 50\%    \\ \hline
	\end{tabular}
	\caption{Performance of different models on CIFAR-10}
\end{table}

To investigate whether our method truly achieves unlearning or approximates the effect of unlearning, we will employ three approaches to study the model. The first approach is based on our observation that, although contrastive models do not require labels during the training process, the output probabilities obtained using their encoders to predict data exhibit high prediction confidence. This difference in prediction confidence is one of the core components in supervised membership inference attacks. Thus, it suggests that an overfitted self-supervised contrastive learning model can also be targeted by supervised membership inference attacks. Our implementation confirms this, with the inference success rate being similar to that of the EncoderMI method. However, our primary focus here is on prediction confidence.

The difference in prediction confidence between training and non-training data is mainly manifested in the model's prediction probability distribution for non-training data, which tends to be relatively flat and uniform. In contrast, the prediction probability distribution for training data exhibits an extremely skewed distribution. The following figure illustrates the distribution of the top-3 prediction confidence values in three-dimensional space for an overfitted contrastive learning model before and after unlearning, with respect to training and non-training data.

\begin{figure*}[!htb]
	\centering
	\begin{minipage}{.333\textwidth}
		\centering
		\includegraphics[width=1\textwidth]{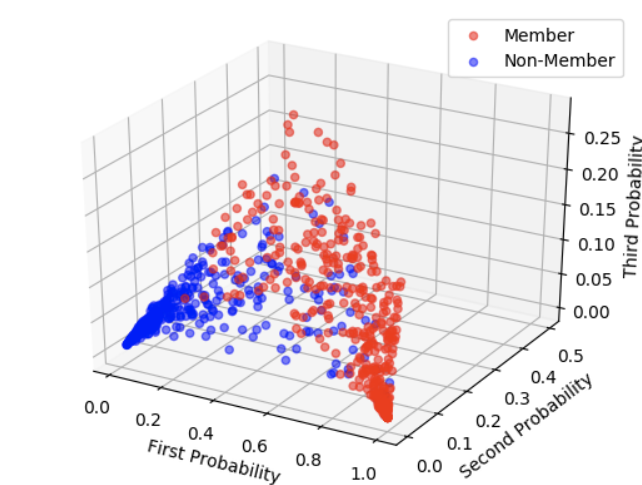}	

	\end{minipage}\hfill
	\begin{minipage}{.333\textwidth}
		\centering
		\includegraphics[width=1\textwidth]{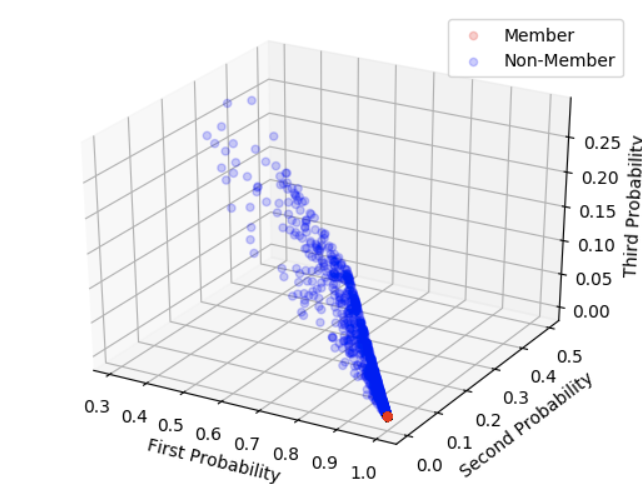}		
	\end{minipage}\hfill
	\begin{minipage}{.333\textwidth}
		\centering
		\includegraphics[width=1\textwidth]{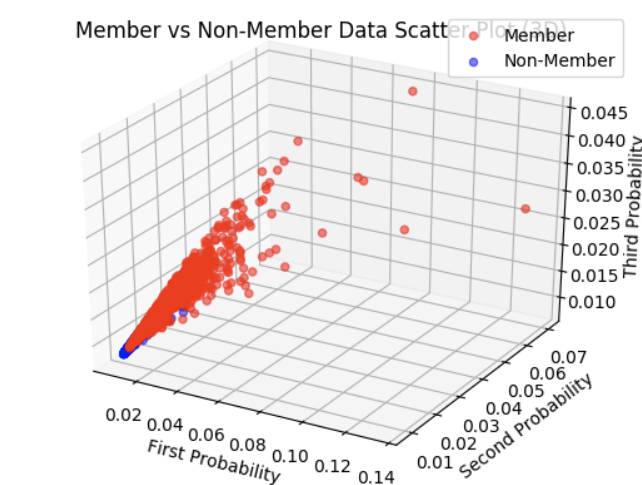}		
	\end{minipage}
	Before Unlearning
\end{figure*}

\begin{figure*}[!htb]
	\centering
	\begin{minipage}{.333\textwidth}
		\centering
		\includegraphics[width=\textwidth]{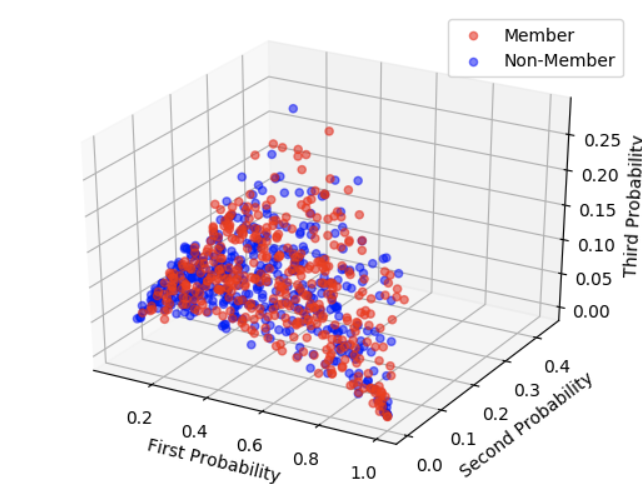}	
		
	\end{minipage}\hfill
	\begin{minipage}{.333\textwidth}
		\centering
		\includegraphics[width=\textwidth]{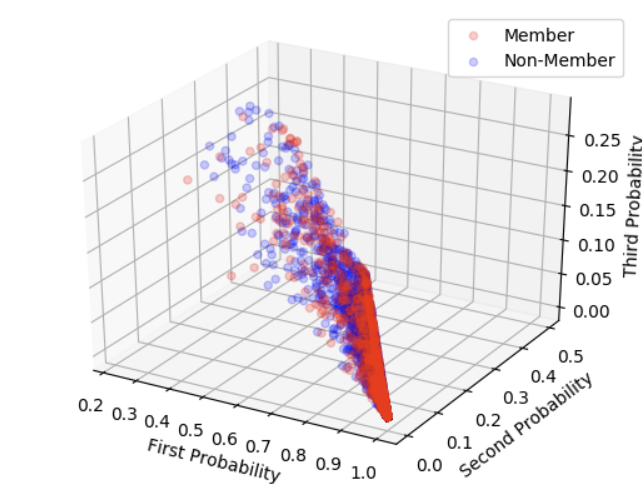}		
	\end{minipage}\hfill
	\begin{minipage}{.333\textwidth}
		\centering
		\includegraphics[width=\textwidth]{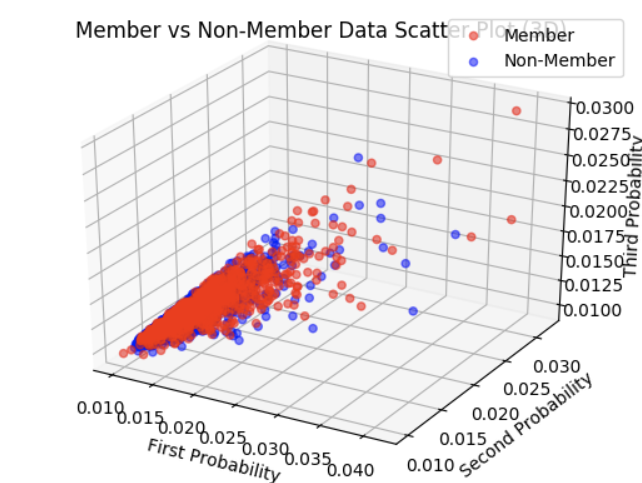}		
	\end{minipage}
	After Unlearning
	\caption{The graph represents the shape of the distribution of the predicted probabilities of the data for models with different degrees of overfitting}
	\label{fig:unlearn}
\end{figure*}

There are various forms of overfitting, and as shown in the figure above, all three distributions of overfitting have close success rates of inference attacks on their members, but their distributions of top3 probability values for team-trained and non-trained data are quite different. The explanation we give here is that the model will have a large variation in its prediction ability or prediction confidence as the degree of overfitting changes. We believe that this may be due to a decrease in the generalization ability of the model and the fact that the internal parameters of the model are tuned to fit the training data more easily as the team training data are trained in depth. We found that when a smaller learning rate is used, the overfitting distribution of the model changes in the direction of the most lateral graph above. When we increase the learning rate, the overfitting distribution of the model will be as shown in the middle of the figure above. As we continue to increase the learning rate, the model overfitting distribution becomes like the rightmost position in the upper panel.Using our forgetting method works for all three types of unlearning, i.e., making the distribution of the training data vary as if it were the distribution of the non-training data, making it indistinguishable

However, this approach can only demonstrate the approximate unlearning effect of our method from one perspective. Next, we will use the cosine similarity method to further explore the changes in the model before and after unlearning. In the self-supervised contrastive learning framework, the model obtains two augmented data samples by applying two different augmentations to the data, which are then used as positive samples for cosine similarity, while the other original data samples in the batch serve as negative samples. The second approach is based on this principle: after a contrastive learning model becomes overfitted, the cosine similarity between the positive samples of the trained data will be very close to 1, whereas the cosine similarity between the positive samples of the non-training data will not be as close to 1. The following figure shows the changes in cosine similarity between the positive samples of member and non-member data before and after unlearning. From the figure, we can observe that after implementing our unlearning method, the cosine similarity values between the positive samples of member and non-member data become almost indistinguishable. This suggests that the unlearning method indeed renders the training data similar to non-member data.

\begin{figure}[!htb]
	\centering
	\includegraphics[width=0.5\textwidth]{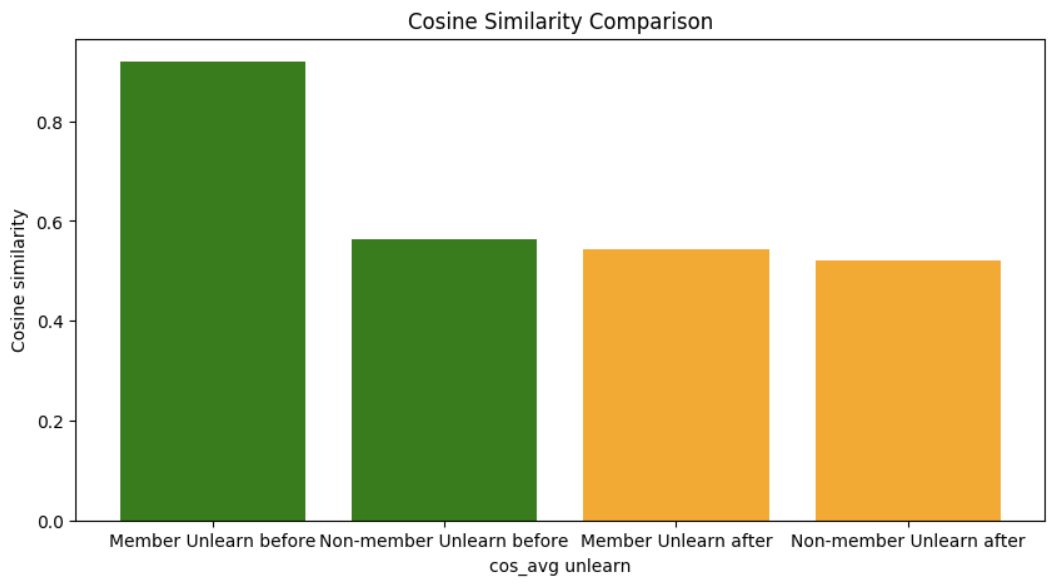}
	\caption{The graph represents the change in cosine similarity between the training and non-training data in the later stages before performing machine unlearning}
	\label{fig:cos_sim}
\end{figure}

The third approach involves comparing the loss differences between training and non-training data. In the contrastive learning framework, the computation of a data sample's loss value is dependent on the other data samples in the batch, as they provide negative sample support. Despite this, calculating the loss for each data sample can still, to some extent, reflect the differences between training and non-training data. We can compute the loss for a batch of member data and a batch of non-member data using the model. The resulting loss differences can reveal the discrepancies between training and non-training data. The following figure illustrates the changes in the loss values for training and non-training data before and after unlearning.

\begin{figure}[!htb]
	\centering
	\begin{minipage}{.5\textwidth}
		\centering
		\includegraphics[width=0.8\textwidth]{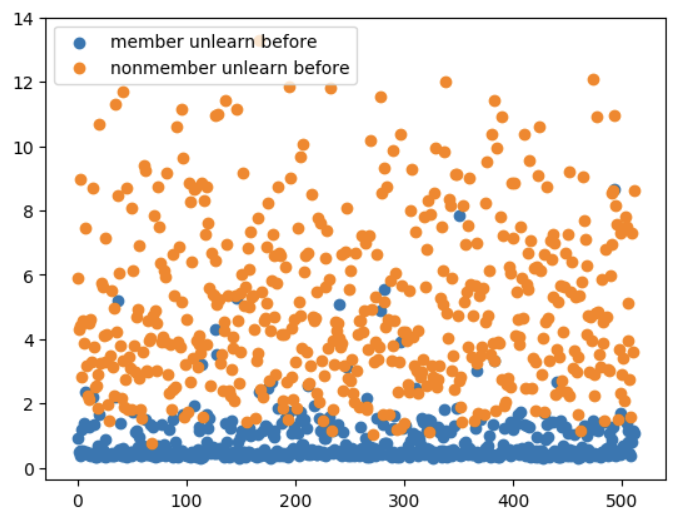}	
	\end{minipage}\hfill
	\begin{minipage}{.5\textwidth}
		\centering
		\includegraphics[width=0.8\textwidth]{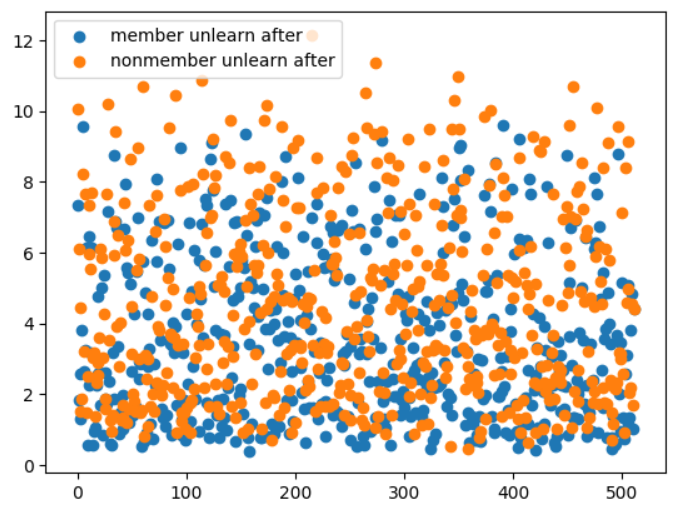}		
	\end{minipage}
	
	\caption{This figure shows the change in loss of a batch of training and non-training data before and after performing machine unlearning}
	\label{fig:data_loss}
\end{figure}

To observe the changes in the model's outputs for training and non-training data before and after unlearning more intuitively, we integrated the three dimensions mentioned above into a three-dimensional feature space. The three features are the model's top-1 prediction probability value for the data, the cosine similarity, and the training loss. The following figure presents the three-dimensional space plot generated by combining these dimensions before and after unlearning. As can be seen from the figure, the original training data exhibits a more concentrated distribution, as training reduces their uncertainty; whereas the non-training data's distribution is more dispersed and random. After executing the unlearning process, the distribution of the original training data becomes as dispersed and random as that of the non-training data.

\begin{figure}[!htb]
	\centering
	\begin{minipage}{.5\textwidth}
		\centering
		\includegraphics[width=0.8\textwidth]{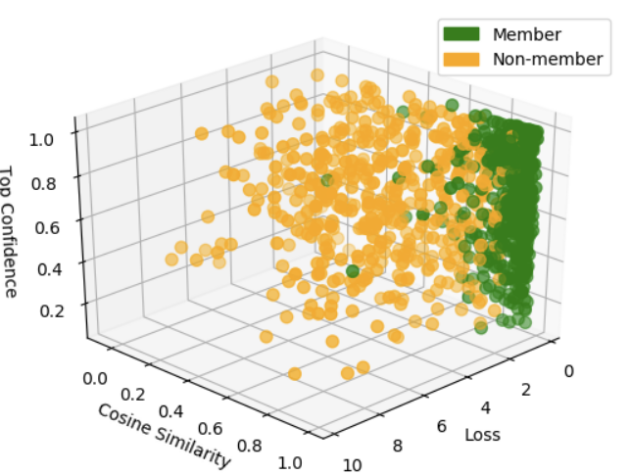}	
	\end{minipage}\hfill
	\begin{minipage}{.5\textwidth}
		\centering
		\includegraphics[width=0.8\textwidth]{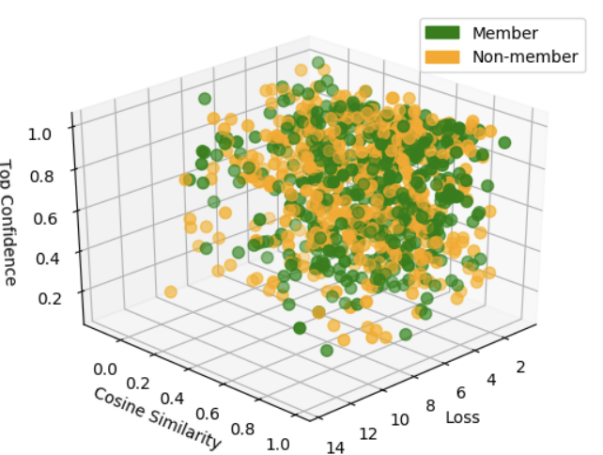}		
	\end{minipage}
	
	\caption{After using our unlearning method, the training data becomes as dispersed in space and highly differentiated from the non-training data}
	\label{fig:feature_combin}
\end{figure}

Our method was initially implemented by modifying the model's training loss, which consists of three components. To demonstrate that all three components play a role in the unlearning process, we carried out an ablation study by breaking down these components and experimenting with various combinations and individual training scenarios to observe the actual contribution of each part during training. The table below presents the results of training with each part individually and in combination with the others. The primary function of mnorm is to reduce the model's prediction confidence; however, it does not alleviate overfitting or render membership inference attacks ineffective. GP is the core component of the loss, capable of disabling membership inference attacks. As seen in the table, using GP alone achieves reasonably good results, and when combined with the model's training loss for the data, it enables disabling membership inference attacks and making member data similar to non-member data with minimal accuracy loss.

\begin{table*}[!htb]
	\centering
	
	\begin{tabular}{|c|c|c|c|c|c|c|c|c|c|c|c|c|}
		\hline
		MNorm & GP & MLoss & \multicolumn{5}{c|}{Before Unlearning} & \multicolumn{5}{c|}{After Unlearning} \\
		\cline{4-13}
		& & & ACC & MIA & TOP1 & LOSS & COSSIM & ACC & MIA & TOP1 & LOSS & COSSIM \\
		\hline
		$\surd$ & $\surd$ & $\surd$ & 70\% & 87\% & 0.98 & 0.8 & 0.92 & 60\% & 51\% & 0.22 & 3.7 & 0.59 \\
		\hline
		$\surd$ & $\surd$ & $\times$ & 70\% & 87\% & 0.98 & 0.8 & 0.92 & 54\% & 50\% & 0.03 & 4.4 & 0.55 \\
		\hline
		$\surd$ & $\times$ & $\times$ & 70\% & 87\% & 0.98 & 0.8 & 0.92 & 10\% & 50\% & nan & nan & nan \\
		\hline
		$\times$ & $\surd$ & $\surd$ & 70\% & 87\% & 0.98 & 0.8 & 0.92 & 62\% & 50\% & 0.4 & 3.2 & 0.59 \\
		\hline
		$\surd$ & $\times$ & $\surd$ & 70\% & 87\% & 0.98 & 0.8 & 0.92 & 57\% & 94\% & 0.35 & 1.23 & 0.92 \\
		\hline
		$\times$ & $\surd$ & $\times$ & 70\% & 87\% & 0.98 & 0.8 & 0.92 & 55\% & 50\% & 0.4 & 1.19 & 0.59 \\
		\hline
	\end{tabular}
	\caption{Ablation study: the effect of different factors on model performance before and after Unlearning}
\end{table*}

In the table, it may seem that mnorm is not important, but in fact, in contrast learning using its trained encoder in the downstream task in order to improve the generalization ability of the model, it is necessary to appropriately reduce its prediction confidence. To satisfy the definition of unlearning, i.e., to make the training data achieve similar output as the non-training data, it is necessary to make the prediction confidence and distribution of the training data to be close to the non-training data. The effect of the simplified version of the latter optimization we did the ablation experiment, the only loss of the method is the gradient penalty value calculated using the training data plus the training loss of the model on the member data, the experimental results are shown in the following table.

\begin{table*}[!htb]
	\centering
	
	\begin{tabular}{|c|c|c|c|c|c|c|c|c|c|c|c|}
		\hline
		MemGP & MLoss & \multicolumn{5}{c|}{Before Unlearning} & \multicolumn{5}{c|}{After Unlearning} \\
		\cline{3-12}
		& & ACC & MIA & TOP1 & LOSS & COSSIM & ACC & MIA & TOP1 & LOSS & COSSIM \\
		\hline
		$\surd$ & $\surd$ & 70\% & 87\% & 0.98 & 0.8 & 0.92 & 60\% & 50\% & 0.25 & 3.9 & 0.57 \\
		\hline
		$\surd$ & $\times$ & 70\% & 87\% & 0.98 & 0.8 & 0.92 & 51\% & 50\% & 0.23 & 4.9 & 0.54 \\
		\hline
	\end{tabular}
	\caption{Ablation study Method two: the effect of different factors on model performance before and after Unlearning}
\end{table*}

To evaluate the performance of our method when unlearning different amounts of data, we conducted a comparative experiment. Under the same conditions, we only adjusted the amount of data to be forgotten and observed the performance changes in the model after unlearning. The experimental results are shown in the table below. As can be seen, the performance of the model after unlearning gradually improves as the amount of forgotten data increases, while still ensuring the failure of membership inference attacks. However, as more data is forgotten, the model's prediction confidence and cosine similarity for member data also gradually increase. To explain this phenomenon, we conducted another experiment. Initially, the model's unlearning training was set to 10 epochs; we reduced the unlearning training epochs to 10 when unlearning 10,000 data samples and found that the model's performance after training was almost identical to that when unlearning 2,000 data samples, as shown in the table. Subsequently, we observed the losses of the two components of the loss during the model's unlearning training in real-time and discovered that once the GP loss decreased to nearly zero, the model could achieve approximate unlearning, rendering membership inference attacks ineffective and making the prediction confidence and cosine similarity of training data similar to those of non-training data.

\begin{table*}[!htb]
	\centering
	
	\begin{tabular}{|c|c|c|c|c|c|c|c|c|c|c|}
		\hline
		Unlearn  & \multicolumn{5}{c|}{Before Unlearning} & \multicolumn{5}{c|}{After Unlearning} \\
		\cline{2-11}
		DataNum
		& ACC & MIA & Top1prob & LOSS & COSSIM & ACC & MIA & Top1prob & LOSS & COSSIM \\
		\hline
		1000  & 70\% & 87\% & 0.98 & 0.8 & 0.92 & 56\% & 51\% & 0.37 & 4.2 & 0.59 \\
		\hline
		2000  & 70\% & 87\% & 0.98 & 0.8 & 0.92 & 60\% & 51\% & 0.25 & 3.9 & 0.57 \\
		\hline
		3000  & 70\% & 87\% & 0.98 & 0.8 & 0.92 & 61\% & 52\% & 0.31 & 3.4 & 0.6 \\
		\hline
		4000  & 70\% & 87\% & 0.98 & 0.8 & 0.92 & 63\% & 52\% & 0.36 & 3.3 & 0.63 \\
		\hline
		5000  & 70\% & 87\% & 0.98 & 0.8 & 0.92 & 64\% & 50\% & 0.4  & 3.2 & 0.65 \\
		\hline
		10000 & 70\% & 87\% & 0.98 & 0.8 & 0.92 & 67\% & 53\% & 0.5  & 2.8 & 0.68 \\
		\hline
	\end{tabular}
	\caption{Model performance before and after unlearning for different amounts of unlearned data}
\end{table*}

As mentioned earlier, our approach can also produce favorable results for supervised learning models. Therefore, we conducted experiments on supervised learning models, using ResNet18 as the experimental model. The evaluation metric for these experiments was the accuracy of the membership inference attack customized for supervised learning, with the prediction confidence probability distribution serving as a secondary evaluation metric. The following table shows the performance of different data sets before and after unlearning using the supervised learning model and retraining for comparison, model accuracy, membership inference attack accuracy, and prediction confidence for trained and untrained data. It can be seen that our method also shows good performance on the supervised learning model.

\begin{table*}[!htb]
	\centering
	
	\begin{tabular}{|c|c|c|c|c|c|c|c|c|c|c|c|}
		\hline
		Dataset & \multicolumn{4}{c|}{Before Unlearning} & \multicolumn{4}{c|}{After Unlearning} & \multicolumn{3}{c|}{Retraining} \\
		\cline{2-12}
		& ACC & MIA & MTop1 & NMTop1  & ACC & MIA & MTop1 & NMTop1  & ACC & MTop1 & NMTop1 \\
		\hline
		cifar10 & 66\% & 80\% & 0.989 & 0.889  & 63\% & 50\% & 0.901 & 0.892  & 62\% & 0.916 & 0.891 \\
		\hline
		cifar100 & 32\% & 90\% & 0.997 & 0.933  & 28\% & 50\% & 0.944 & 0.935  & 29\% & 0.944 & 0.935 \\
		\hline
		svhn & 90\% & 70\% & 0.977 & 0.912  & 88\% & 50\% & 0.885 & 0.881  & 85\% & 0.892 & 0.874 \\
		\hline
	\end{tabular}
	\caption{Supervised model performance before, after Unlearning and during Retraining for different datasets}
\end{table*}

We employed the t-SNE dimensionality reduction technique to observe the changes in the model's prediction outputs for the data before and after unlearning, as illustrated in the figure below. The results clearly show that after applying our unlearning method, the prediction outputs for training and non-training data become increasingly similar. Before unlearning, the distribution of training data is highly concentrated, with relatively large distances between different classes, indicating that the model exhibits high certainty in its predictions for the training data. However, after executing the unlearning operation, we observed a significant increase in the prediction uncertainty for each class, causing the distances between classes to become closer. This outcome indicates that our method successfully treats training data as though it had never been trained, thereby protecting the data effectively. This treatment makes it difficult to distinguish between training and non-training data, effectively preventing membership inference attacks. Through this experiment, we have demonstrated the effectiveness of our method in protecting the privacy of training data and provided strong support for further research in machine unlearning.

\begin{figure}[!htb]
	\centering
	\begin{minipage}{.5\textwidth}
		\centering
		\includegraphics[width=0.8\textwidth]{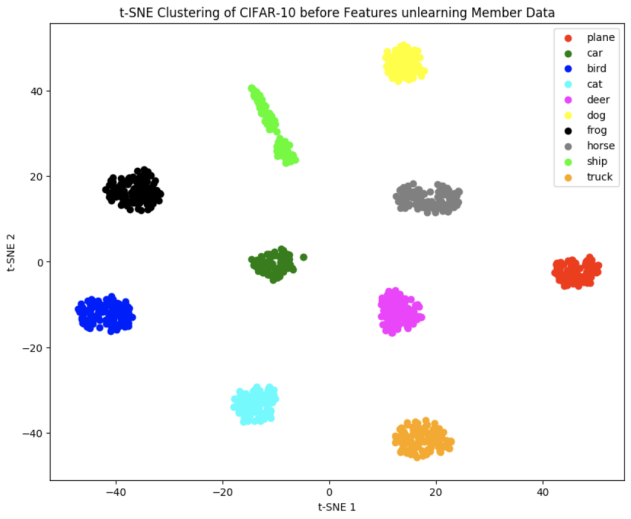}	
	\end{minipage}\hfill
	\begin{minipage}{.5\textwidth}
		\centering
		\includegraphics[width=0.8\textwidth]{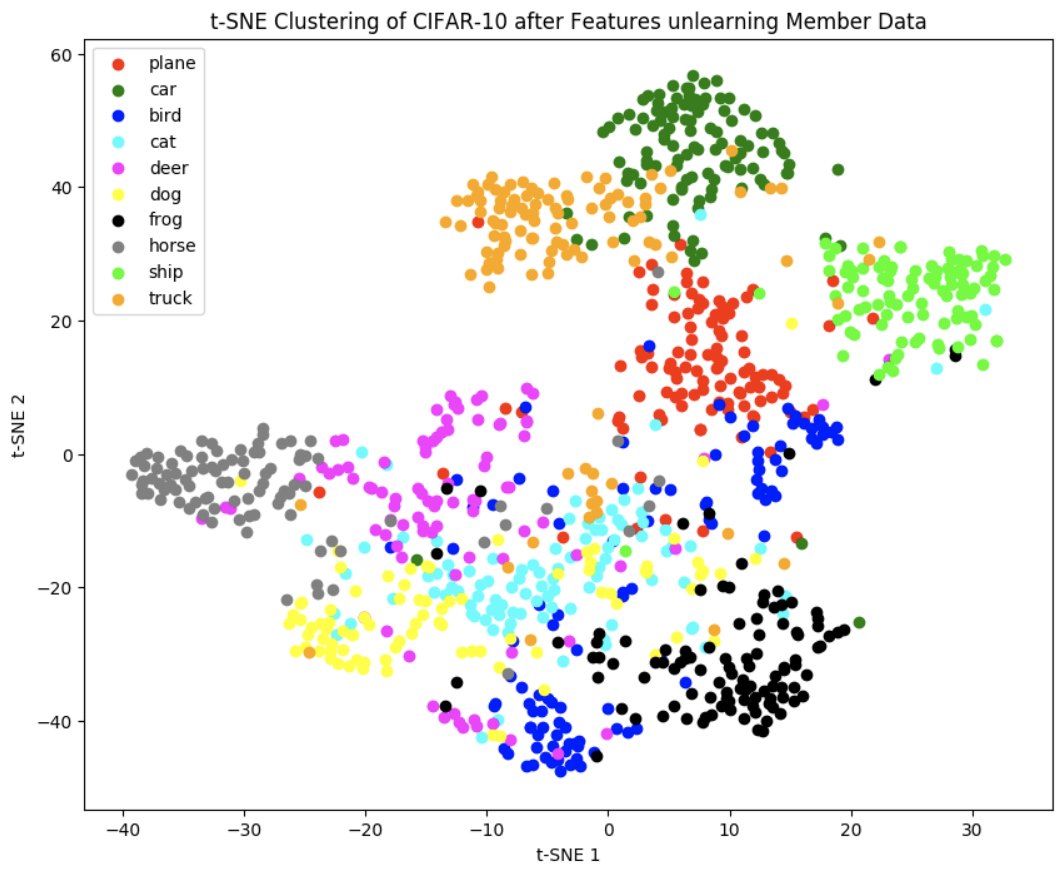}		
	\end{minipage}
\end{figure}
\begin{figure}[!htb]
	\centering
	\begin{minipage}{.5\textwidth}
		\centering
		\includegraphics[width=0.8\textwidth]{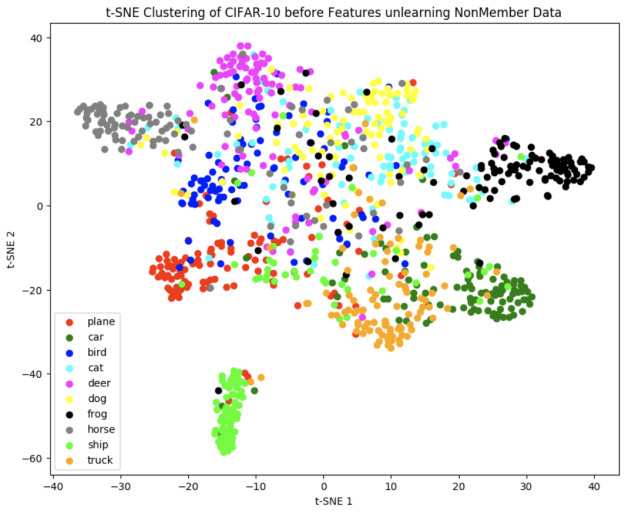}	
	\end{minipage}\hfill
	\begin{minipage}{.5\textwidth}
		\centering
		\includegraphics[width=0.8\textwidth]{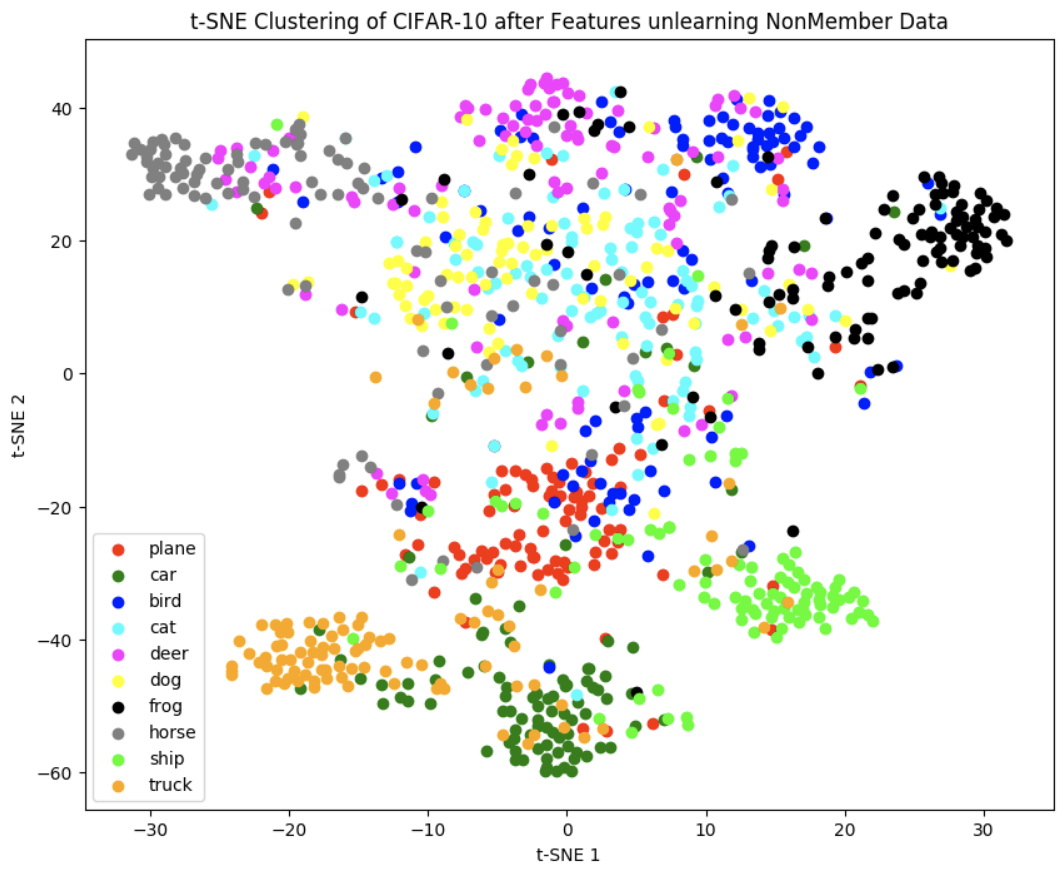}		
	\end{minipage}
	\caption{The graph represents the same t-sne dimensionality reduction to observe the distribution of the training and non-training data in each category before and after machine unlearning}
	\label{fig:t-sne}
\end{figure}

\section{Conclusion}
We propose a machine unlearning method that supports both contrastive learning models and supervised models, achieving excellent performance levels. Our approach effectively defends against membership inference attacks (MIAs) and protects user privacy. Moreover, it does not require complex preprocessing, nor does it rely on specific frameworks, making it a fairly generalizable method. To implement our method, one simply needs the model and the data to be forgotten, making the approach highly user-friendly. Additionally, our method does not demand extensive computational resources; it can be achieved with just a few training epochs. However, further evaluation and testing, such as examining the model's unlearning effects from various perspectives, remain areas for future research.


\end{document}